# DAVID: Influence Diagram Processing System for the Macintosh

Ross D. Shachter
Department of Engineering-Economic Systems
Stanford University
Stanford, CA   94305-4025

Influence diagrams are a directed graph representation for uncertainties as probabilities. The graph distinguishes between those variables which are under the control of a decision maker (decisions, shown as rectangles) and those which are not (chances, shown as ovals), as well as explicitly denoting a goal for solution (value, shown as a rounded rectangle). The figure on the right, shows the influence diagram for the oil wildcatter problem from Raiffa [1968], as it is represented in DAVID. The arcs in the diagram indicate the probabilistic dependence among random variables with respect to a particular factorization of the joint distribution, and also indicate the time (decision point) at which information becomes available.

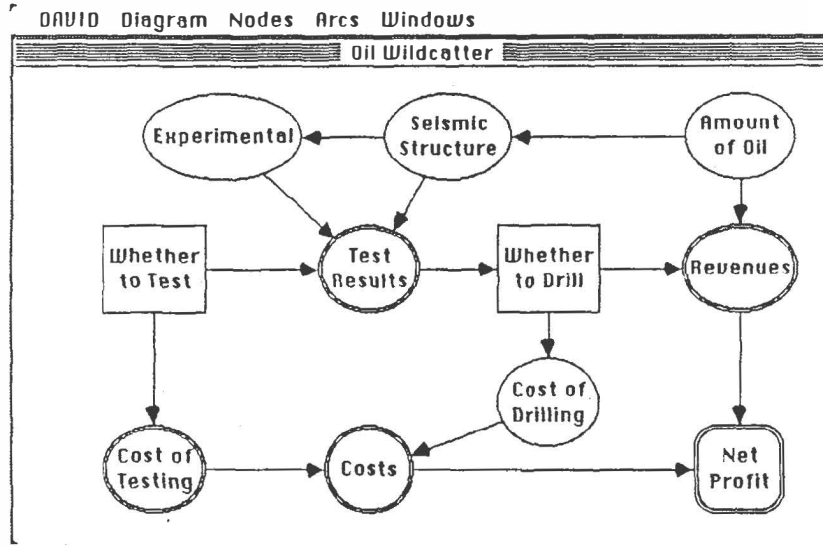

Influence diagrams have been used for the last ten years as a model structuring and elicitation device in the practical field of decision analysis. They have been a powerful communication tool during the initial discussion about a problem, as well as when explaining results after analysis. Because the diagrams are heirarchical, with the numbers "hidden" within the nodes, attention is focused on the relationship among variables and the assumptions of timing and observability. (The figure on the next page shows the "opened" node for seismic structure.) Traditionally, the actual analysis has been performed using other data structures, principally trees, and the problem has been converted from its assessed version to be analyzed.

Within the last few years, a number of theoretical results allow for the analysis to be performed directly on the influence diagram, as assessed. In fact, for many problems, this representation offers computational advantages, since it explicitly captures conditional independence among variables. (For example, the experimental test results are conditionally independent of the amount of oil, given the seismic structure.) Another benefit is in the calculation of sensitivities, wherein we consider changes to the original problem structure. In particular, we can easily adjust the informational assumptions in a problem, to determine the relative value of observing variables at different times. (For example, we could see the value of knowing the amount of oil at the time of our drilling decision by adding an arc from amount of oil to drill.) In general, these techniques apply a sequence of transformations to different influence diagrams, to solve either probabilistic inference or decision analysis problems.

The latest version in the recent series of software efforts to manipulate influence diagrams is the DAVID program on the Macintosh. It is written in LISP (ExperLisp) so it is fairly transportable and easy to wrap into a shell within an expert system. The focus, however, is on the use of graphical interaction in the construction, manipulation and analysis of influence diagram models. The system is a working demonstration of the ability of people to think about models within a probabilistic framework. Many of the criticisms of probabilistic models of uncertainty are overcome by an intelligent graphical interface that explicitly incorporates conditional independence. The figure above shows how the conditional distribution for seismic structure is entered into DAVID. By storing the model how it is initially formulated and allowing graphical modification, DAVID encourages analysts to reassess and experiment.

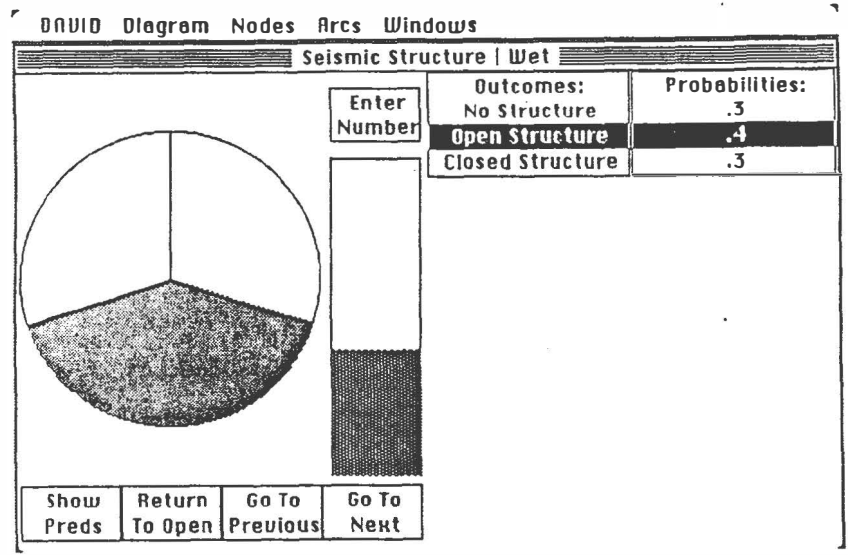

A problem is evaluated in DAVID by "reducing" nodes through a series of value-preserving transformations, as shown on the right. Fundamentally, these operations are conditional expectation, maximization of expected utility and the application of Bayes' Theorem. In the influence diagram, these appear as the removal of a chance node, removal of a decision node, and the reversal of an arc between chance nodes, respectively (Howard and Matheson [1968], Olmsted [1983], and Shachter [1984, 1986]). Any completely specified influence diagram can be solved using these operations. At the end, the decision nodes are replaced by deterministic "policies," showing the optimal decisions given the information available at the time of the decision. If we look inside the value node after the problem is reduced (shown on the next page), then we can see the optimal value of objective function.

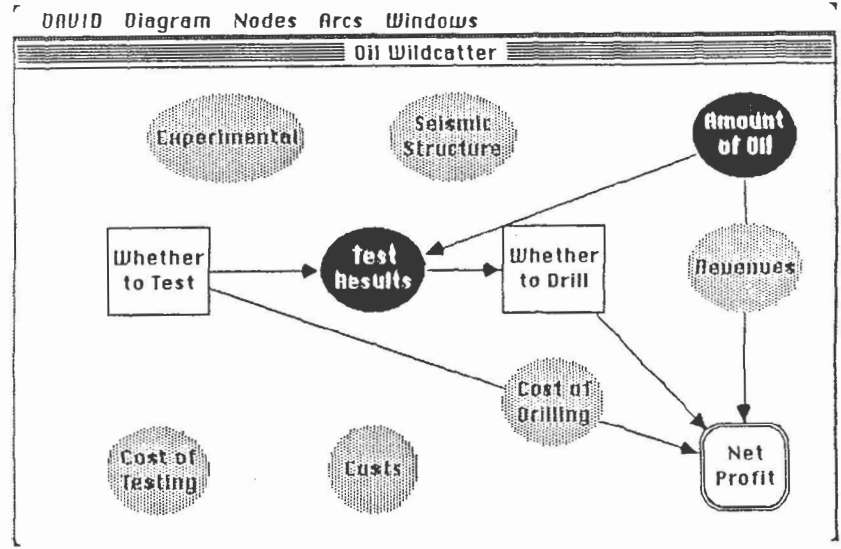

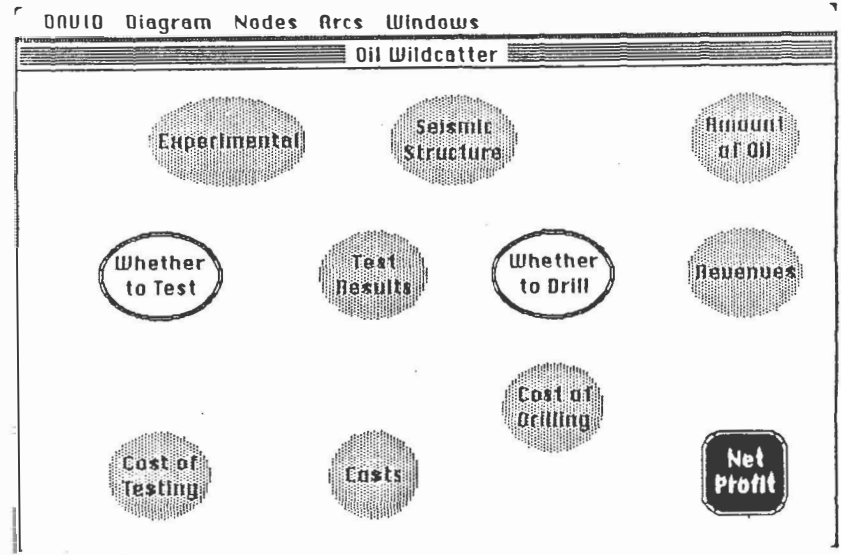

245

Another benefit of the reduction process is that every intermediate product is a valid influence diagram, so one can think of the reduction process as "consolidating" the information in the model. DAVID takes advantage of this property to increase the power available to the user willing to think in terms of influence diagrams. For example, it is straightforward to obtain the value lottery for the optimal policy in the oil wildcatter problem, shown on the right.

One of the goals of the DAVID project is to develop an environment in which people can be comfortable thinking in probabilistic models. While DAVID is a prototype implementation of an influence diagram processor, the results have been most encouraging. Not only have students been able to quickly build and solve decision models with DAVID, but it has stimulated their interest and understanding of influence diagram and decision theory as well.

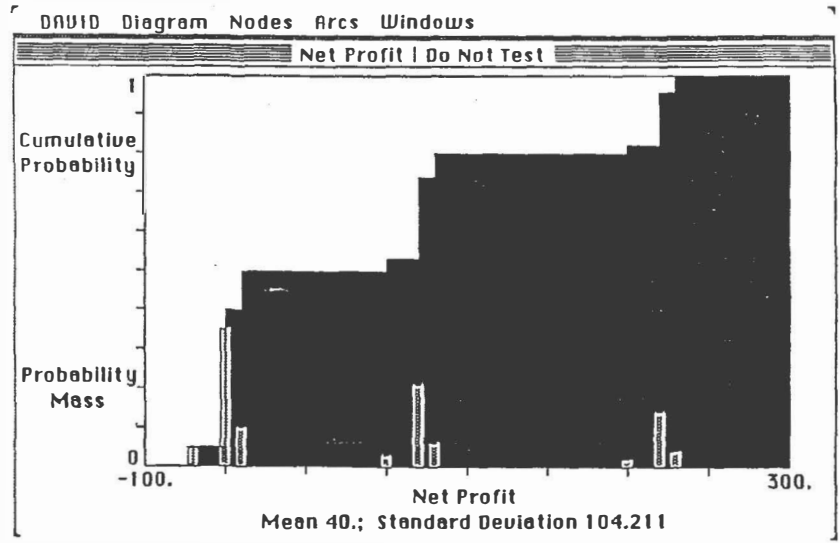

There are many features and conveniences built into the current prototype DAVID program which there is no room to illustrate here, and more that will be available soon in a followon version. For example, the paradigm is able to exploit the principle of optimality and dynamic programming whenever possible (Tatman [1985]), such as in the equipment replacement problem shown in the figure on the next page.

There are many exciting possibilities using influence diagrams as a representation for communication among people and machines, and as a language for the development of expert systems. Because DAVID is written in LISP, there are a variety of expert system shells available from which it can be invoked.

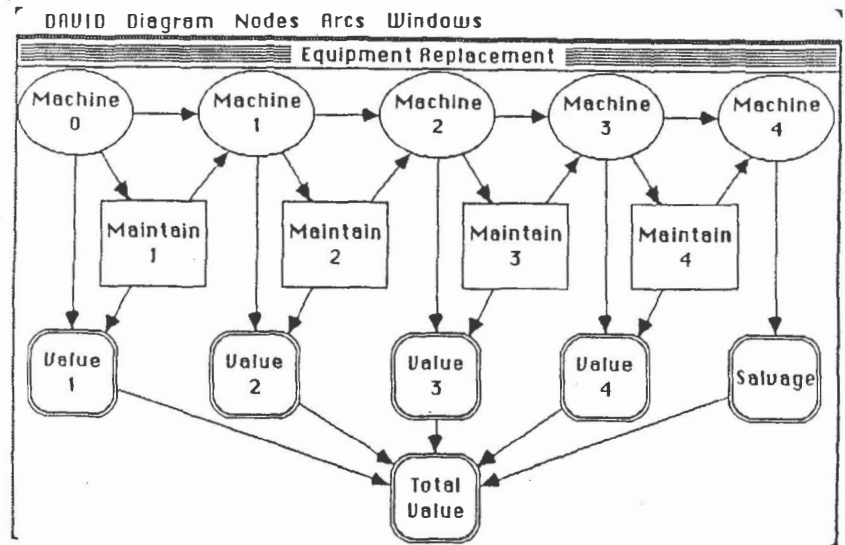

The latest version of the program will be available for demonstration and possible distribution at the workshop.

## References

Howard R. A., and J. E. Matheson. 1981. *Influence Diagrams,* in Howard, R. A. and J. E. Matheson, eds., 1984, The Principles and Applications of Decision Analysis, Vol. II, Strategic Decisions Group, Menlo Park, CA.

Olmsted, S. M. 1984. *On Representing and Solving Decision Problems,* Ph.D. Thesis, EES Department, Stanford University.

Raiffa, H. A. 1968. Decision Analysis, Addison-Wesley, Reading, MA.

Shachter, R. D. 1984. *Evaluating Influence Diagrams,* to appear in Operations Research.

Shachter, R. D. 1986. *Probabilistic Inference and Influence Diagrams,* EES Department, Stanford University.

Tatman, J. A. 1985. *Decision Processes in Influence Diagrams: Formulation and Analysis,* Ph.D. Thesis, EES Department, Stanford University.

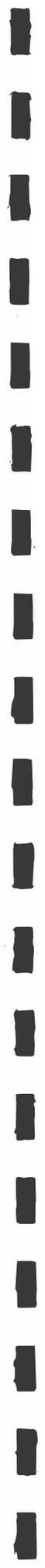